%% file: tmlr.tex
\documentclass[10pt]{article} 
\usepackage[preprint]{tmlr}

\input{math_commands.tex}

\usepackage{hyperref}
\usepackage{url}
\usepackage{amsthm}
\usepackage{graphicx}
\usepackage[frozencache=true,cachedir=minted-cache]{minted}
\usepackage{placeins}

\newtheorem{theorem}{Theorem}

\title{A Simple Method for PMF Estimation on Large Supports}

\author{\name Alex Shtoff \email alexander.shtoff@tii.ae \\ Technology Innovation Institute\\}

\newminted[pycode]{python}{tabsize=4, fontsize=\scriptsize, frame=lines, framesep=\fboxsep, rulecolor=\color{gray!40}}

\def\month{MM}  
\def\year{YYYY} 
\def\openreview{\url{https://openreview.net/forum?id=XXXX}} 

\DeclareMathOperator{\diag}{diag}

\begin{document}

\maketitle

\begin{abstract}
We study nonparametric estimation of a probability mass function (PMF) on  $[N] = \{0, 1,2, \dots, N - 1\}$ from samples when $N$ can be large and the true PMF is often multi-modal and heavy-tailed. The core idea is to treat the empirical PMF as a signal on a line graph and apply a data-dependent low-pass filter. Concretely, we form a symmetric tri-diagonal operator $\mH = \mL - \diag(\vp)$ - the path graph Laplacian $\mL$ plus a diagonal perturbation built from the empirical frequency vector $\vp$, then compute the first $k$ eigenvectors, corresponding to the smallest eigenvalues. Projecting the empirical PMF onto this $k$-dimensional subspace produces a smooth, multi-modal estimate that preserves coarse structure while suppressing noise. A light post-processing step of clipping and re-normalizing yields a valid PMF. 

Because $\mH$ is symmetric tridiagonal, the computation is reliable and runs in $O(kN)$ time and requires $O(kN)$ memory for a PMF over $[N]$ using standard symmetric tridiagonal eigensolvers. We also provide a practical, data-driven rule for selecting $k$ based on an orthogonal-series risk estimate, so the method “just works’’ with minimal tuning. On synthetic and real heavy-tailed examples, the approach preserves coarse structure while suppressing sampling noise, compares favorably to logspline and Gaussian-KDE baselines in the intended regimes. However, it has known failure modes (e.g., abrupt discontinuities). The method is short to implement, robust across sample sizes, and suitable for automated pipelines and exploratory analysis at scale because of its reliability and speed.
\end{abstract}

\section{Introduction}
Estimating a densities from empirical observations is a fundamental problem with applications ranging from visualizing a histogram to complex generative models for images and text. In this work we focus on an extremely simple and fundamental instance - estimating the PMF of a discrete random variable supported on $[N] = \{0, 1, 2, \ldots, N - 1\}$ from empirical observations. Our primary focus is on a large $N$, between thousands and a few millions, where the PMF being estimated is multi-modal and ``heavy tailed''. Due to the finite support, we obviously are not referring to heavy tailedness in the classical sense, but we do mean that a large portion of mass is spread far away from a small number of areas of concentration.

\begin{figure}[tbph]
    \centering
    \includegraphics[width=.9\textwidth]{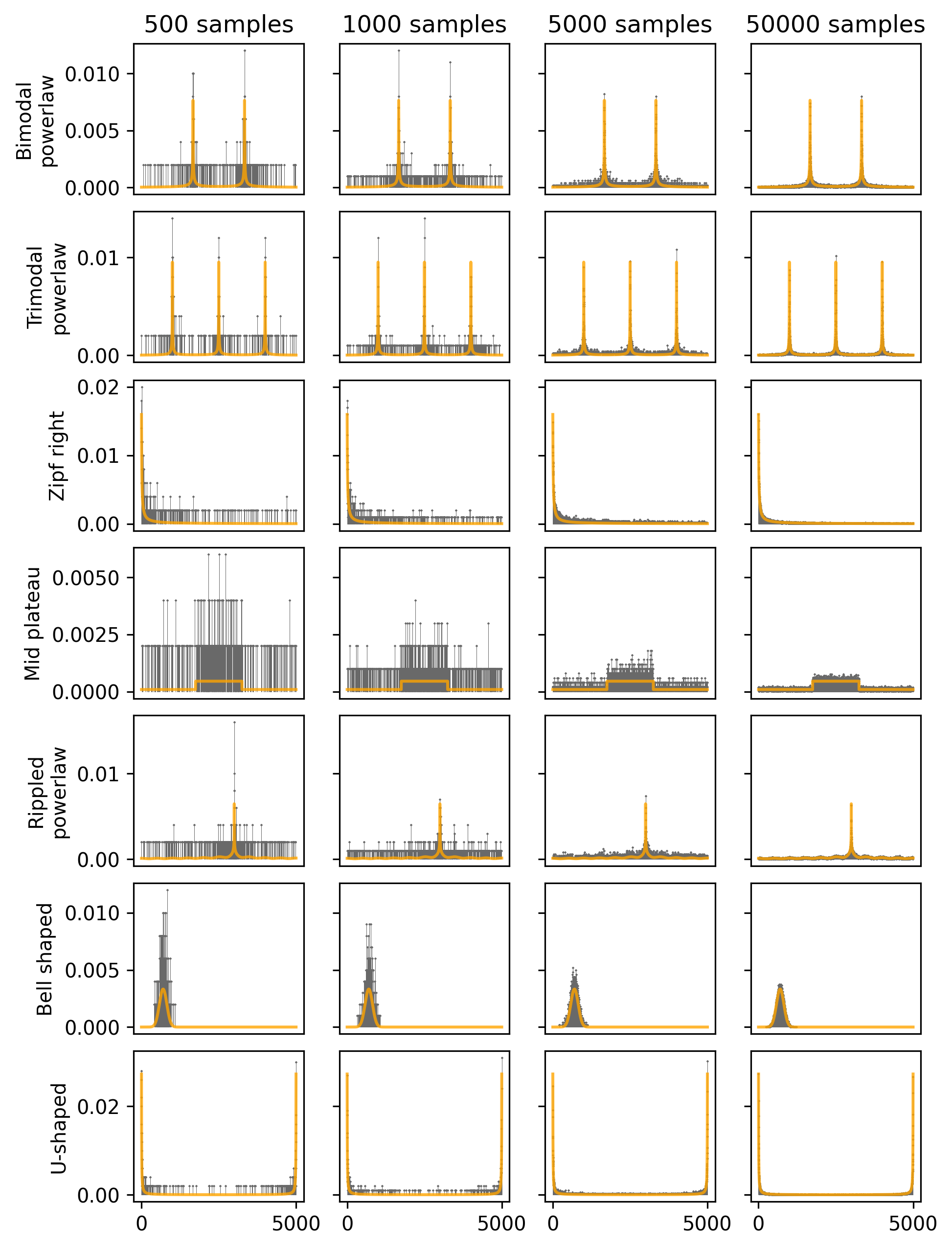}
    \caption{Empirical frequency vectors (gray) and underlying PMFs that generated them (orange).}
    \label{fig:distributions}
\end{figure}

Examples include, but not limited to, the well-known power-law distributions, such as the Zipf distributions $p(n)\propto (a+n)^{-b}$, their centered variants $p(n) \propto (a + |n - \mu|)^{-b}$, and their mixtures. Such distributions can be naturally obtained from counts, such as token counts in a corpus of documents in the language domain, or the number of page visits in the recommender systems domain. High resolution discretizations of continuous naturally heavy-tailed quantities, such as the time since the last user interaction with a given item, product prices, network latencies, or ad auction bids, also yield similar discrete PMFs. The main feature of the empirical frequency vectors stemming from such distributions is that they are form \emph{noisy} and \emph{sparse} vectors with entries spread throughout, as demonstrated by Figure \ref{fig:distributions}. Indeed, it can be seen that the only distribution whose empirical frequency vectors are different is the \emph{bell shaped} distribution from in the figure, whose samples are mostly concentrated around its (light-tailed) bell. These distributions shall serve us throughout the paper to analyze and demonstrate our method's strengths and weaknesses.

To the best of our knowledge, the three most commonly employed univariate density estimators available in typical data analysis software are binned empirical frequencies \citep{binning_pearson},  Gaussian Kernel Density Estimators (KDEs) \citep{kde_parzen,kde_rosenblatt}, and the \emph{logspline} methods \citep{KooperbergStone1991,StoneHansenKooperbergTruong1997} implemented in the popular R package having the same name \citep{logsplineCRAN}. Bins and Gaussian KDEs are extensively used in standard data visualization libraries, such as Plotly \citep{plotly}, Vega-Altair \citep{vega_altair}, and Seaborn \citep{seaborn}. However, for heavy-tailed distributions it's hard to determine the right placement of bins to get both a reasonably accurate and interpretable distribution, whereas with Kernel density estimators one must set the right kernel and its width hyper-parameters, which may be challenging and non-intuitive in this setting. While effective heuristics exist, they fail in the cases we focus on in this paper. As we show in our experiments, logspline does prove to be very effective, but occasionally suffers from numerical difficulties and turns out to be slightly less reliable, and thus less suited for automated pipelines, or dashboarding at scale.

To complement the existing toolbox, we believe that a this fundamental and extremely simple problem deserves an extremely simple solution. Thus, our main contribution is a simple, non-parametric, and fast method that fits such heavy-tailed densities well, and is reliable to compute without tuning or oversight. Our main applications in mind are automated visualization and data pipelines that need to run without human intervention, such as pipelines for feature engineering in continually-learning recommender systems \citep{offset_adaptive}, or continually-updating dashboards. We focus on reducibility to simple building blocks having well-developed algorithms and software components, while working reasonably well. Thus, we occasionally make deliberate design decisions that favor speed, reliability, and simplicity over accuracy. In turn, this yields a reliable method whose implementation is just a few lines of code calling linear algebra routines. 

Despite it being simple to reason about and implement, mathematically the proposed estimator is a complex non-linear function of the empirical histogram. Proving formal guarantees was out of our reach as of the time of writing of this paper, and we make no attempt to do so in the paper. Instead, we believe that its interpretation from a familiar standpoint in machine learning, that of minimizing a balance between regularization and fitting, fosters trust, while its computational efficiency and simplicity eases adoption.

After a short literature review in \Secref{sec:review}, we first present our method's core in \Secref{sec:method}, and demonstrate its strengths and weaknesses with a few synthetic distributions. Then, in \Secref{sec:analysis} we analyze it mathematically, and in \Secref{sec:autotune} we show a heuristic for eliminating the need for manual tuning. Finally, in \Secref{sec:experiments} we compare our method on datasets with heavy-tailed features to Gaussian KDEs with automatically selected bandwidth and the logspline method with default hyperparameters.

\section{Related work}\label{sec:review}

We situate our approach relative to four strands: (i) classical density/histogram smoothing, (ii) Laplacian and spectral methods for smoothing on graphs, (iii) Schr\"odinger-type operators obtained by adding a potential to the Laplacian, and (iv) PMF estimation by orthogonal projections.

\paragraph{Histograms, kernels, and mixtures.} Nonparametric density estimation via kernels goes back to \citet{kde_rosenblatt,kde_parzen}. In practice, both histograms and KDE rely on simple bandwidth/bin-width rules such as those of \citet{Scott1979} and \citet{FreedmanDiaconis1981}, and the averaged shifted histogram provides an inexpensive bridge between histograms and kernels \citep{Scott1985ASH}. A complementary parametric route is finite mixtures, typically fit with the EM algorithm \citep{Dempster1977EM} and surveyed by \citet{McLachlanPeel2000}. . Another widely used alternative is \emph{logspline} density estimation, which models the log-density with cubic splines and selects knots by likelihood-based forward--backward steps; foundational treatments include \citet{KooperbergStone1991,KooperbergStone1992} and the extended spline modeling framework of \citet{StoneHansenKooperbergTruong1997}, with mature implementations such as the R package \emph{logspline} \citep{logsplineCRAN}. Extensions cover binned and spectral settings \citep{Koo2000,KooperbergStoneTruong1995Rate}. These families are effective but introduce tuning - bandwidths, knot numbers/penalties, or component counts and nonconvex fitting - which our goal is to avoid with a parameter-light spectral construction.

\paragraph{Laplacian smoothing and spectral viewpoints.} In computer graphics, Laplacian-based fairing acts as a low-pass smoother \citep{Taubin1995,Desbrun1999ImplicitFairing}. In machine learning, graph-Laplacian regularization underpins manifold learning (Laplacian Eigenmaps, Diffusion Maps), harmonic-field semi-supervised learning, and manifold regularization in RKHS \citep{BelkinNiyogi2003,CoifmanLafon2006,ZhuGhahramaniLafferty2003,BelkinNiyogiSindhwani2006}. A complementary signal-processing perspective advocates spectral filtering and multiscale constructions such as graph wavelets \citep{Shuman2013SPMag,Hammond2011}; for piecewise-smooth targets, $\ell_1$ variants like graph trend filtering provide edge preservation \citep{Wang2016GraphTrendFiltering}. We retain this spectral template but alter resulting bases through a \emph{data-dependent} potential derived from the empirical histogram, which concentrates eigenfunctions near peaks while preserving global regularity.

\paragraph{Laplacians with potentials.} Adding a diagonal potential to the Laplacian yields Schr\"odinger-type operators whose eigenfunctions localize in regions indicated by the potential. \citet{CzajaEhler2013PAMI} introduced \emph{Schr\"odinger Eigenmaps} for semi-supervised learning by incorporating barrier potentials. Our construction instantiates this operator-with-potential principle on a path graph with a potential equal to the empirical histogram, leading to the tridiagonal operator $\mL_{\mathrm{path}} - \diag(\vp)$. Our approach resembles semi-classical signal analysis \citep{semiclassical}, where Schr\"odinger Eigenmaps of a similar nature to ours are used to derive a data dependent basis for signal analysis for pulse-like signals, such as heart-rate. Smoothing is then obtained by truncating its spectrum. Our work can be seen as a direct variant of similar ideas to sparse noisy empirical frequency vectors whose underlying PMF has a pulse-like shape.

\paragraph{Orthogonal projection based estimators} 
Projection estimators that expand a density in an orthonormal system and truncate after a certain number of terms date back at least to \citet{cencov1962estimation}, and were given a concise treatmet by \citet{schwartz_orthogonal_density}, who proved consistency for orthogonal series estimators with sample-average coefficients, and slowly increasing truncation. \citet{kronmal_targer_orth} supplied a clear MISE (mean integrated squared error)  decomposition andpractical truncation rules. \citet{watson_orth_1969} framed these series estimators alongside kernel methods. Subsequent work \citep{diggle_hall_orthogonal_selection,hart_orth} documented failures of the earliest rules for truncating the series on multimodal densities, and proposed a more stable model selection method. Our method is a projection estimator too, but onto a \emph{data-dependent} orthonormal system, rather than one chosen a-priori. However, we do adopt the Diggle-Hall style expected error criterion to choose the truncation level, as a heuristic.

\section{The basic method}\label{sec:method}
To describe our method, we denote by $\Delta_N$ the $N$-dimensional probability simplex, by $[x]_+ = \max(x, 0)$ the \emph{relu} function, and by $\mL$ the \emph{Laplacian} matrix
\begin{equation}\label{eq:laplacian}
\mL =\begin{pmatrix}
    1 & -1 & 0 & 0 & \cdots & 0 \\
    -1 & 2 & -1 & 0 & \cdots & 0 \\
    0 & -1 & 2 & -1 & \cdots & 0 \\
    \vdots & \vdots & \ddots & \ddots & \ddots & \vdots \\
    0 & 0 & \cdots & -1 & 2 & -1 \\
    0 & 0 & \cdots & 0 & -1 & 1
    \end{pmatrix},
\end{equation}
which lies at the heart of our method. Note, that $\mL$ is symmetric and tridiagonal.

Having observed samples from $[N]$, the inputs to our method are the vector $\vp \in \Delta_N$ of the empirical sample frequencies, and a small integer $k \in \sN$, e.g., $k = 10$. The method consists of the following four steps:
\begin{enumerate}
\item Form the symmetric and tri-diagonal matrix $\mH = \mL - \diag(\vp)$.
\item Compute the eigenvectors corresponding to the $k$ \emph{smallest} eigenvalues of $\mH$ in the columns of the matrix $\mV = [\vv_1 \vert \vv_2 \vert \cdots \vert \vv_k]$.
\item Compute $\vu = \mV (\mV^T \vp)$
\item Return the vector $\vq$ with $\evq_i = [u_i]_+ / \|[\vu]_+ \|_1$.
\end{enumerate}
At first glance it may not be obvious why those four steps. That the third step is merely a projection onto the subspace spanned by the eigenvectors, whereas the fourth step is just a normalization. Hence, this concrete subspace is the main pillar of our method's ability to fit sparse histograms, whereas the availability of on efficient tri-diagonal eigenvalue solvers is the main pillar of our method's efficiency and reliability.

To appreciate the method's simplicity, below is a Python implementation:
\begin{pycode}
import numpy as np
import scipy.linalg as la

def estimate_pmf(samples, k=10):
    # compute empirical frequencies
    p = np.bincount(samples, minlength=np.max(samples) + 1)
    p = p.astype(float) / np.sum(p)

    # Form the diagonals of H
    diag_l = np.r_[1, np.full(p.size - 2, 2), 1]  # [1, 2, ..., 2, 1]
    diag_h = diag_l - p
    off_diag_h = np.full(p.size - 1, -1)          # [-1, ..., -1]

    # compute first k eigenvectors of H
    _, v = la.eigh_tridiagonal(diag_h, off_diag_h, select="i", select_range=(0, k - 1))

    # project and normalize
    u = v @ (v.T @ p)
    return np.maximum(u, 0) / np.sum(np.maximum(u, 0))
\end{pycode}

But before taking a closer look, we would like to draw the readers' attention to Figure \ref{fig:apx_synthetic_500}, where we demonstrate using synthetic examples where our method shines versus where it does not. It is apparent that ``spiky'' distributions having significant mass both around their modes but also far away are fit well, whereas Gaussian kernel-density estimation falls short. Our method is slighly less suited to wide bell-shaped distributions, in contrast to Gaussian kernel-density estimation that shines in these cases. Finally, our method performs poorly for discontinuous densities, in contrast to kernel-density estimators that, despite fitting poorly, performs gracefully. Our analysis in the next section attempts to explain these phenomena.

\section{Exploratory analysis}\label{sec:analysis}
Although our method is a straightforward instance of Schr\"odinger eigenmaps, and we could simply cite that literature, we present a direct, elementary analysis to offer clearer insight and broader accessibility. We hope that this will inspire similar techniques to be applied to other domains of analyzing univariate data. We rely only on basic linear algebra facts, and offer the viewpoint of optimizing an objective that balances data fitting and regularization, the bread and butter of modern machine learning. Some steps may be routine for readers familiar with related techniques, but this elementary treatment, in our view, makes the key ideas accessible to the general machine-learning audience.

Recall that any real symmetric matrix always has real eigenvalues, and the corresponding eigenvectors form an orthonormal set. A well-known characterization of the eigenvalues and eigenvectors of symmetric matrices is via minimization of quadratic functions. Formally, we have:
\begin{theorem}\label{thm:courant_fischer_orth}
Let $\mA$ be a real symmetric $n \times n$ matrix, let $\lambda_1 < \dots  < \lambda_n$ be its eigenvalues, and let $\ve_1, \dots, \ve_n$ be the corresponding orthonormal eigenvectors. Then,
\[
\ve_k = \argmin_{\vx} \{ \vx^T \mA \vx : \|\vx\|_2 = 1,~\vx^T \ve_1 = 0,~\ldots,~\vx^T \ve_{k-1} = 0 \}.
\]
\end{theorem}
In other words, the eigenvector corresponding to the $k$-th smallest eigenvalue is the unit-norm minimizer the quadratic function $\vx^T \mA \vx$ that is orthogonal to all previous eigenvectors. Even though it is not stated in this form in the literature, this result commonly appears \emph{as a part of} proving the famous Courant-Fischer theorem. See, e.g., the proof of Theorem 4.2.8 in \citet{horn2012matrix}. 

To elucidate our method, we shall apply the theorem to the matrix $\mH = \mL - \diag(\vp)$, where $\mL$ is the Laplacian defined in \eqref{eq:laplacian}, and $\vp$ is the vector of empirical frequencies. Note, that our use of Theorem \ref{thm:courant_fischer_orth} requires the eigenvalues to be distinct. Fortunately, any symmetric tri-diagonal matrix with nonzero off-diagonal entries, including our matrix $\mH$, has distinct eigenvalues (e.g., Lemma 7.7.1 in \citet{parlett1998symmetric}).

So by Theorem \ref{thm:courant_fischer_orth}, the eigenvector corresponding to its smallest eigenvalue minimizes the function
\[
\vx^T (\mL - \diag(\vp)) \vx = \vx^T \mL \vx - \vx^T \diag(\vp) \vx = \underbrace{\vx^T \mL \vx}_{(*)} - \underbrace{\langle \vx^2, \vp \rangle}_{(**)},
\]
where $\vx^2$ denotes component-wise squaring. It is easy to see that
\[
(*) = \vx^T \mL \vx = \sum_{i=1}^{N-1} (\evx_i - \evx_{i-1})^2,
\]
and therefore this is a penalty for $\vx$ being non-smooth. The term $(**)$ is the dot-product similarity between $\vx^2$ and $\vp$, and thus it is a reward for $\evx_i$ having a large magnitude whenever $\evp_i$ is large. 

The first eigenvector allocates its unit-norm mass in a manner that minimizes the balance between these two terms, and thus we expect it to yield a smooth sequence that captures the overall shape of the density. The second eigenvector aims to minimize the same balance while being orthogonal to the first; thus, it may allocate its mass to capture additional details about the shape while being smooth. This pattern repeats for subsequent eigenvectors, each capturing additional details that the previous ones have not. Consequently, the first $k$ eigenvectors corresponding to the smallest eigenvalues form a dictionary of orthogonal atoms that are adapted to the shape of the distribution.

\begin{figure}[htbp]
    \centering
    \includegraphics[width=.8\textwidth]{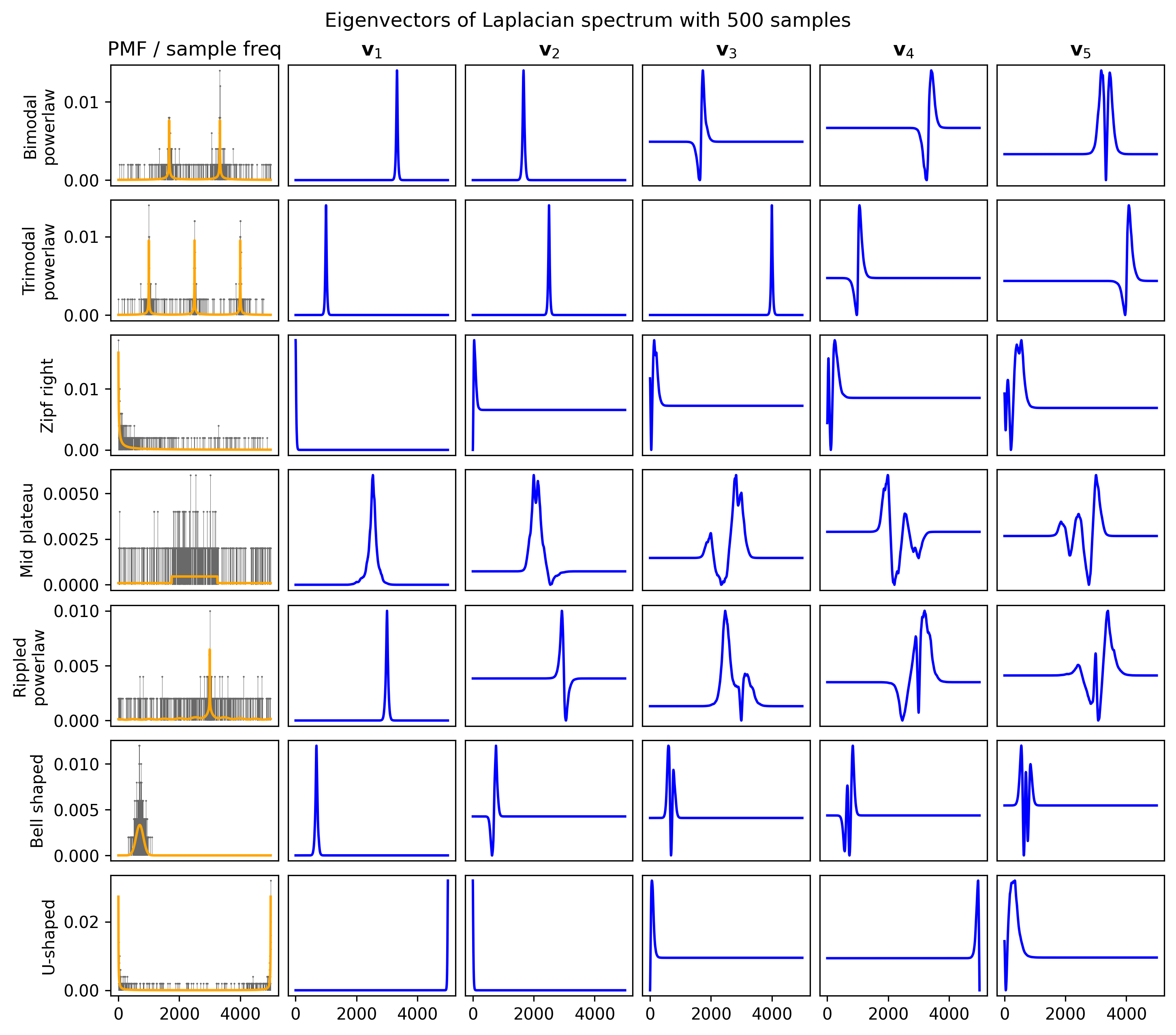}
    \caption{PMFs, samples, and eigenvectors. The left-most column has the true PMF supported on $0, 1, \dots, 999$ in orange, and empirical frequencies $\vp$ obtained from 500 samples in blue. The next columns depict the first five eigenvectors of $\mL - \diag(\vp)$.}
    \label{fig:eigenvectors}
\end{figure}

\Figref{fig:eigenvectors} demonstrates the eigenvectors obtained from samples coming from our catalog of synthetic distributions. We can observe empirically that in a multi-modal distribution, the first eigenvector typically captures the most dominant mode: the data alignment term draws its mass towards the mode, whereas the smoothness penalty makes it discard other modes. Having captured all dominant modes, the remaining eigenvectors appear to oscillate with increasing frequency, with oscillations mostly around the modes.

This phenomenon is not a coincidence. The eigenvectors of the Laplacian matrix $\mL$ itself are the celebrated Discrete Cosine Transform \citep{the_dct}, which are renowned for their ability to approximate smooth sequences with just a few vectors. In contrast to the discrete cosine transform, which is a data-independent basis, the eigenvectors $\vv_1, \dots, \vv_k$ of our perturbed Laplacian are a \emph{data-dependent} basis that concentrates its approximation power in a way that is aligned with the shape of the distribution being fit.

\begin{figure}[tbhp]
    \centering
    \includegraphics[width=.8\textwidth]{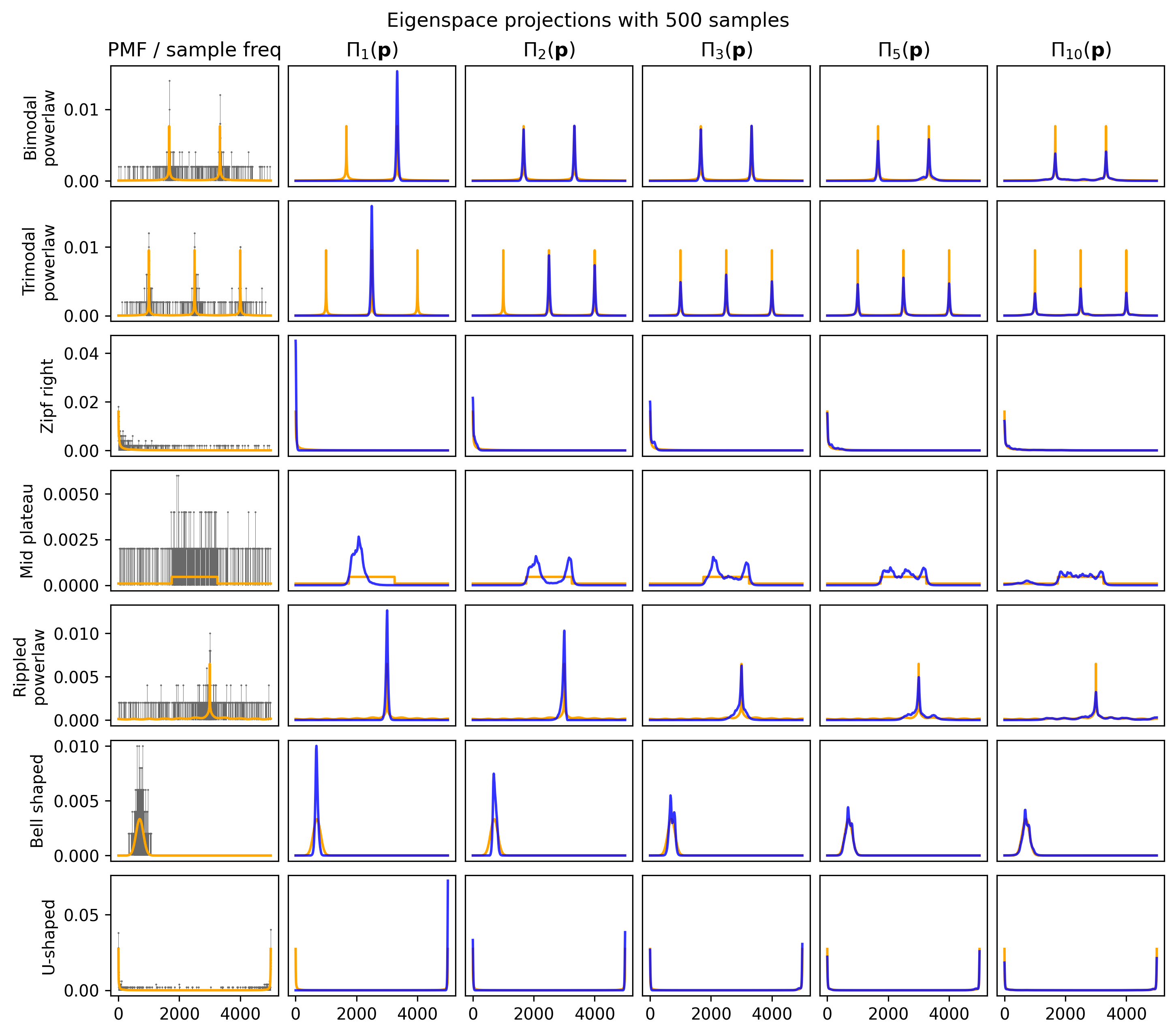}
    \caption{The projections of the empirical frequencies stemming from 500 samples from $\{0, 1, \dots, 4999\}$ onto the first $k$ eigenvectors of $\mL - \diag(\vp)$ for different values of $k$. Rows are various distribution shapes. The first column depicts true and empirical PMFs. The next columns are the projections of this empirical PMF onto eigenspaces of increasing dimension, depicted in blue.}
    \label{fig:apx_synthetic_500}
\end{figure}

\Figref{fig:apx_synthetic_500} continues \figref{fig:eigenvectors}, and depicts the projections of the empirical frequencies onto the space spanned by the first $k$ eigenvectors, for various values of $k$ with 500 samples. As is apparent, the spiky shape of the heavy-tailed mixtures is nicely captured by the projection. In contrast, a bell-shaped distribution is poorly captured, even with 10 eigenvectors. A distribution with an abruptly changing PMF, the ``Mid plateau'', is a case when our method fails.

\begin{figure}[tbhp]
    \centering
    \includegraphics[width=.8\textwidth]{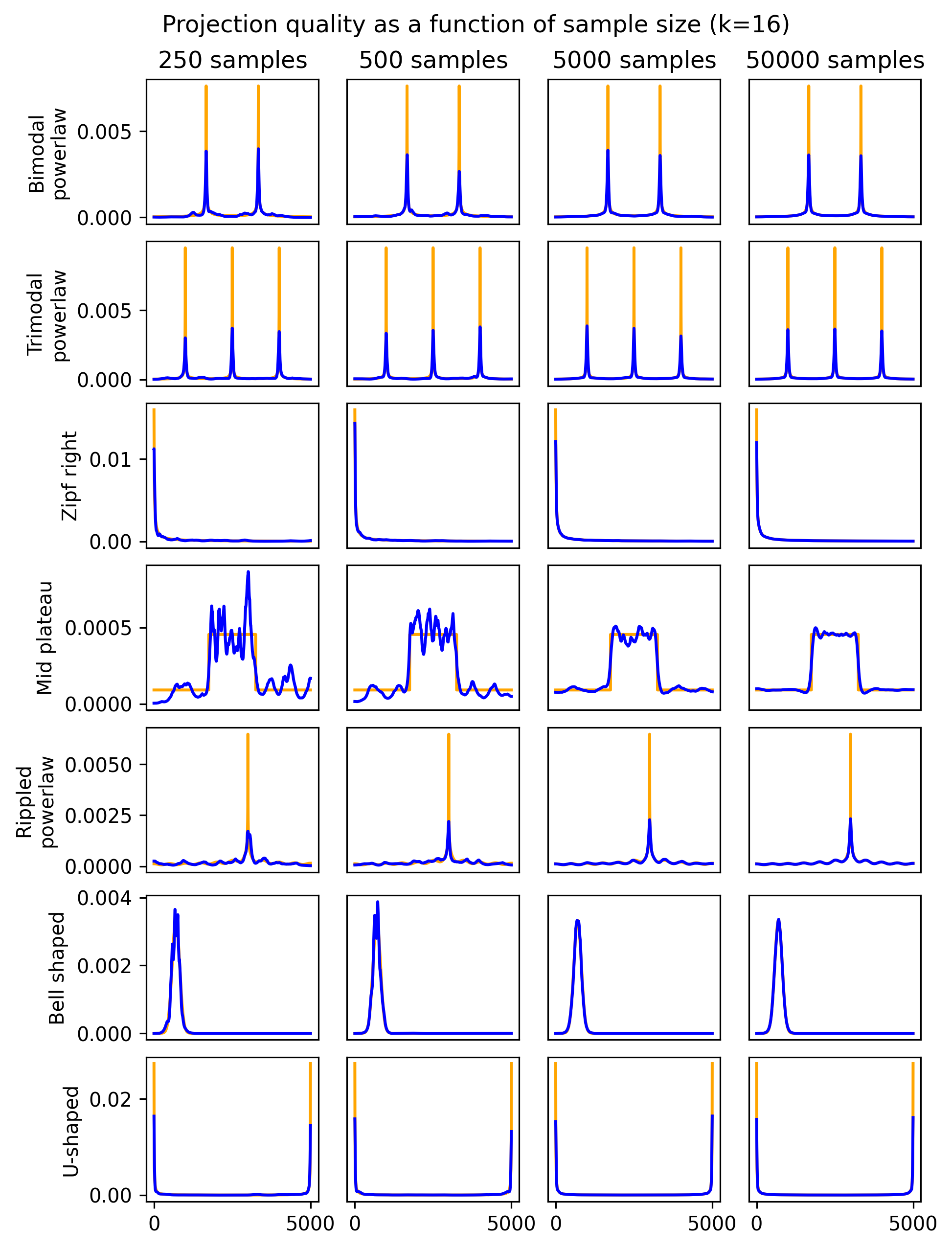}
    \caption{The projections of the empirical histograms stemming from  samples from $\{0, 1, \dots, 4999\}$ of various sizes onto the first $k=16$ eigenvectors of $\mL - \diag(\vp)$. Rows are various distribution shapes. Columns represent sample sizes. Our method in blue, whereas the true density in yellow.}
    \label{fig:apx_sample_sizes}
\end{figure}

To appreciate how our method scales with various sample sizes, we plot the projection onto the first $k=16$ eigenvectors the empirical frequency vectors coming from increasingly large samples in \figref{fig:apx_sample_sizes}. The same behavior persists - the shape of spiky distributions is nicely captured even at small sample sizes. A bell-shaped distribution requires more samples to capture. And finally, our method is poorly suited in the case of an abruptly changing PMF, the ``Mid plateau''. 

\section{Automated selection of the number of eigenvalues}\label{sec:autotune}
Since the entire spectrum of the matrix is an orthonormal basis, it can exactly represent the empirical frequency vector. Thus, using the subspace of just the first $k$ first eigenvectors is a form of regularization, where less eigenvectors correspond to stronger regularization. Intuitively, the more data we have, the less we may need to regularize. Indeed, looking again at \figref{fig:apx_sample_sizes} we can see that 16 eigenvectors appear to be too much for small samples. Thus, we need some technique to select the appropriate number $k$ of eigenvectors to use. Such a selection technique, if turns out to work well, also eliminates the need to specify $k$ and makes our method easier to use in practice.

We adopt a heuristic similar to \emph{logspline} described in \citet{StoneHansenKooperbergTruong1997} to select an upper bound on $k$. Then, we select the best $k$ based on a criterion similar to the Bayesian Information Criterion (BIC) \citep{schwarz1978estimating}. We do realize that cross-validation, and other techniques may yield more accurate PMF approximations, but again - we favor simplicity and speed over squeezing every ounce of approximation quality as a design decision, and demonstrate that it works reasonably well.

Concretely, our upper bound on the number of eigenvectors is identical to the \emph{logspline} upper bound on the number of spline knots, $K = \min(4 n^{1/5}, n / 4, N, 30)$, where $n$ is the number of observations in the sample, and $N$ is the number of unique values in the sample. For selecting the number of spline knots, logspline uses the Bayesian Information Criterion (BIC). 

Since our fitting procedure is a \emph{Euclidean projection onto an orthogonal basis}, rather than log-likelihood minimization, we cannot use BIC directly for estimating the expected error. Rather, we use the variant described in \citet{diggle_hall_orthogonal_selection}: given an orthonormal basis $\mV = [\vv_1 \lvert \dots \rvert \vv_K]$ and an empirical PMF vector $\vp$ obtained from $n$ observations, first we compute
\[
\vc = \mV^T \vp, \quad \vs^2 = (\mV \odot \mV)^T \vp, \quad \bar{\vc}^2 = \frac{1}{n-1} \left[n \vc^2 -  \vs^2\right]_+,
\]
where $\odot$ is the Hadamard product, and use these to estimate the expected L2 error estimate of using just $m$ out of $K$ eigenvectors:
\[
\mathcal{E}(m) = \frac{1}{n}\sum_{k=1}^m (s_k^2 - \bar{c}_k^2) + \sum_{k=m+1}^K \bar{c}_k^2.
\]
Consequently, we use $k_{\mathrm{best}} = \argmin_{1 \leq m \leq K} \mathcal{E}(m)$ eigenvectors. Note, that the vectors $\vc$, $\vs^2$, and $\bar{\vc}^2$ are computed \emph{once}, and then $k$ is selected based on these vectors. Explaining rationale behind this formula is beyond the scope of this paper, but we refer readers to \citet{diggle_hall_orthogonal_selection} for its derivation. Even with the above mechanism for automatically selecting $k$, the entire algorithm fits in a short Pythons snippet that can be found in Appendix \ref{apx:estimator}. 

In \figref{fig:performance_comparison_synthetic} we compare the algorithm algorithm to logspline\footnote{We use version 	2.1.22 of logspline. This is the latest stable version as of the time of the writing of this paper.}, and Gaussian kernel-density estimation as availabe in SciPy \citep{scipy} with automatic bandwidth selection. Our objective is observing fitting quality \emph{without} human intervention or tuning. We can see that our method indeed performs well in the heavy-tailed setting, as does logspline. Our method is competitive with logspline, each having its advantages in different settings and different sample sizes. Out of these examples, Gaussian kernel density estimation fits well only in the bell-shaped case. As predicted, our method does \emph{not} perform well on the ``Mid plateau'' PMF, in contrast to both logspline and kernel density estimation, that despite low fitting quality, perform quite gracefully.

\begin{figure}[tbhp]
    \centering
    \includegraphics[width=.8\textwidth]{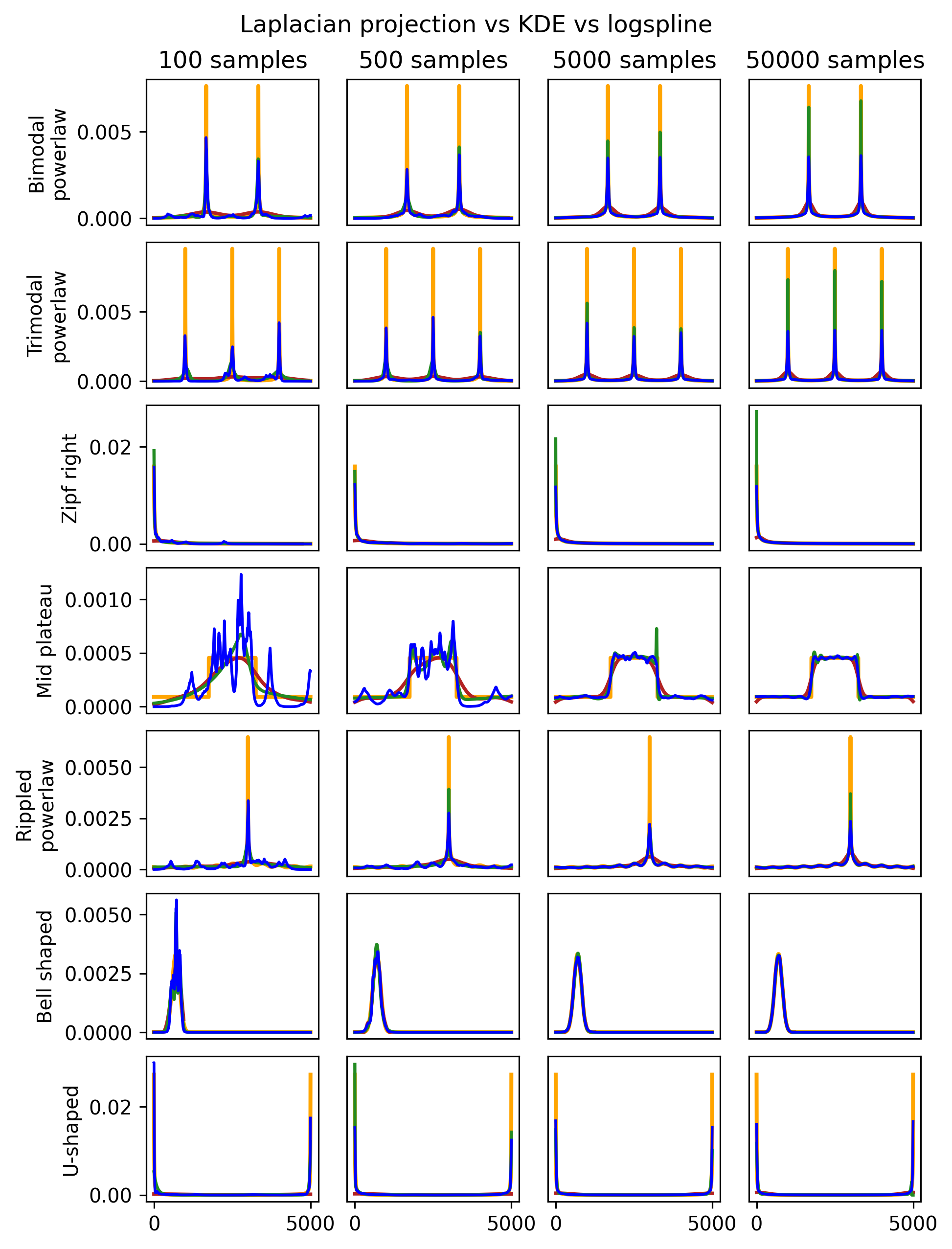}
    \caption{Fits obtained for different sample sizes with out method (blue), Gaussian kernel-density estimate (red), and logspline (green). The true PMF is in yellow. Rows are different distributions, whereas columns are different sample sizes.}
    \label{fig:performance_comparison_synthetic}
\end{figure}

The above error estimate, despite being just an upper bound, also implicitly assumes that the basis is chosen a-priori, independent of the data. This is not the case in our method - the columns of $\mV$ heavily depend on the data by construction of the eigen-problem. Thus, this method is just a heuristic. Viable alternatives include a split of the samples to train and validation sets, or even the use of cross-validation. But again,  we make a conscious design decision of using a method we know to be just a heuristic, while gaining simplicity and speed.As we just demonstrated on the synthetic data, and in the next section on real data - this turns out to work reasonably well in practice.

\section{Testing on real data}\label{sec:experiments}
We use the \emph{spambase} dataset \citep{spambase_94} for spam classification, due to a large number of heavy-tailed numerical columns representing token frequencies, letting us test our method in different settings. All frequencies as multiplied by $1000$ and rounded, to obtain integers as our method expects. We also use the bank marketing dataset \citep{bank_marketing_222} that has only \emph{one} numerical column, the bank account balance, but has a relatively large number of rows that let us test our method with various sample sizes.

Practical real-world numerical features tend to be both \emph{zero-inflated} and \emph{heavy tailed}. This is also the case with our two data-sets. We assume the zero inflation is dealt with separately, by estimating the probability of each feature being zero, and obtaining a separate PMF estimate assuming it is nonzero. Thus, for both datasets, we discard zero samples and test our methods on the non-zero samples.

\begin{figure}[tbh]
    \centering
    \includegraphics[width=.85\textwidth]{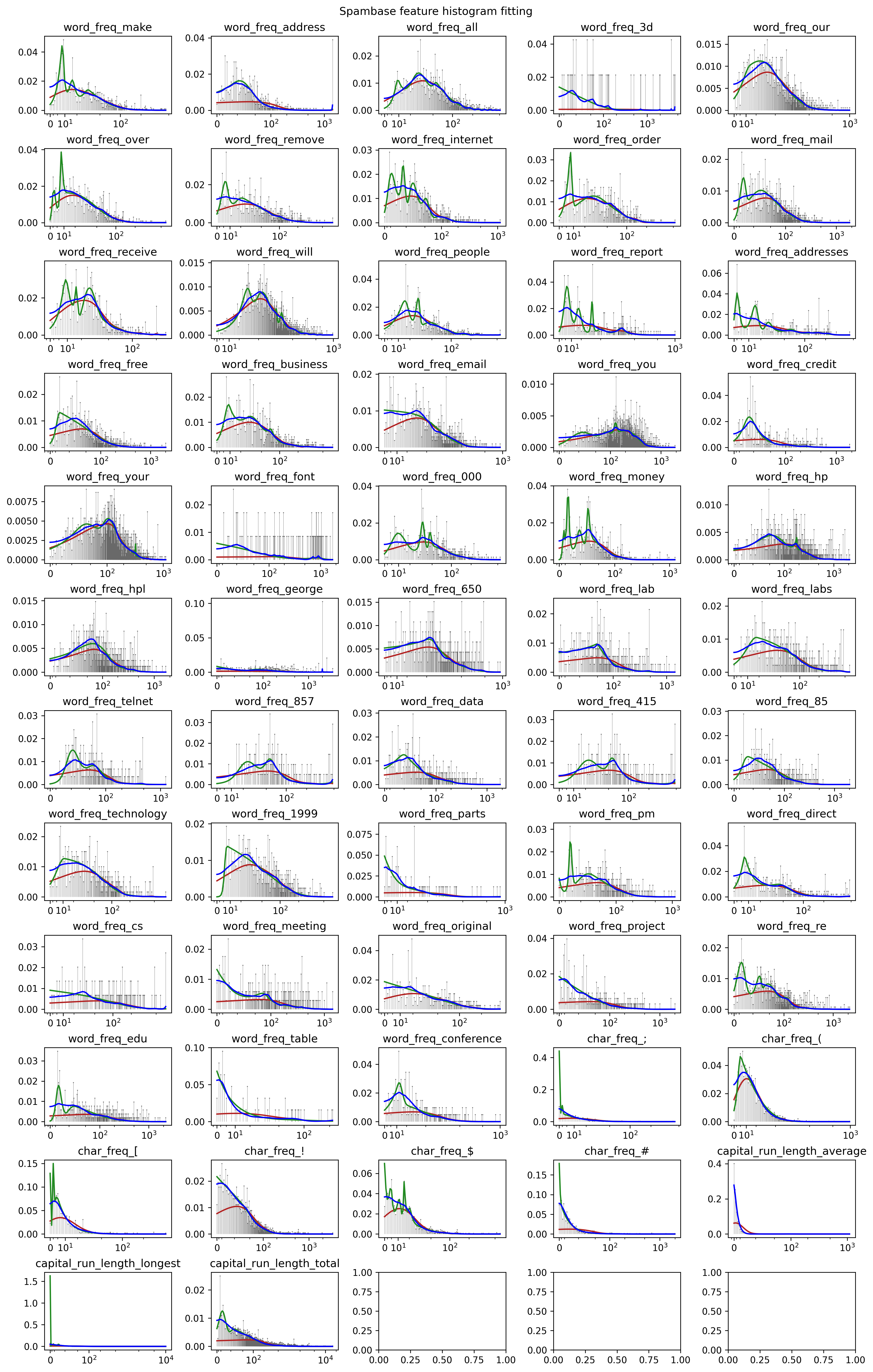}
    \caption{Empirical frequencies (gray) and PMF estimates for columns of the spambase dataset. Gaussian kernel-density estimator in red, logpline in green, and our method in blue.}
    \label{fig:spambase}
\end{figure}

In \figref{fig:spambase} we plot PMF estimates for all the feature columns in the data-set using our method (with automated selection of $k$), logspline, and Gaussian kernel density estimator. Apparently, Gaussian KDE fails to produce a reasonable estimate for many columns, such as the \texttt{word\_freq\_report}, \texttt{word\_freq\_free}, and others. logspline and our method are quite competitive, but slightly different.  We can also see that for some features logspline produced somewhat unreasonable estimates, such as the \texttt{char\_freq\_;}, \texttt{char\_freq\_(}, and \texttt{char\_freq\_\#} features. 

One interesting observation is that for  one feature is missing a logspline estimate altogether -  \texttt{capital\_run\_length\_average} feature. This is because logspline \emph{failed to converge}, and threw an exception with convergence failure as its message. It turns out that failure to converge, even if not common, still may happen. While for manual exploratory data analysis this is not an issue in practice, for automated data pipelines or frequently updating dashboards it is.

\begin{figure}[tbh]
    \centering
    \includegraphics[width=\textwidth]{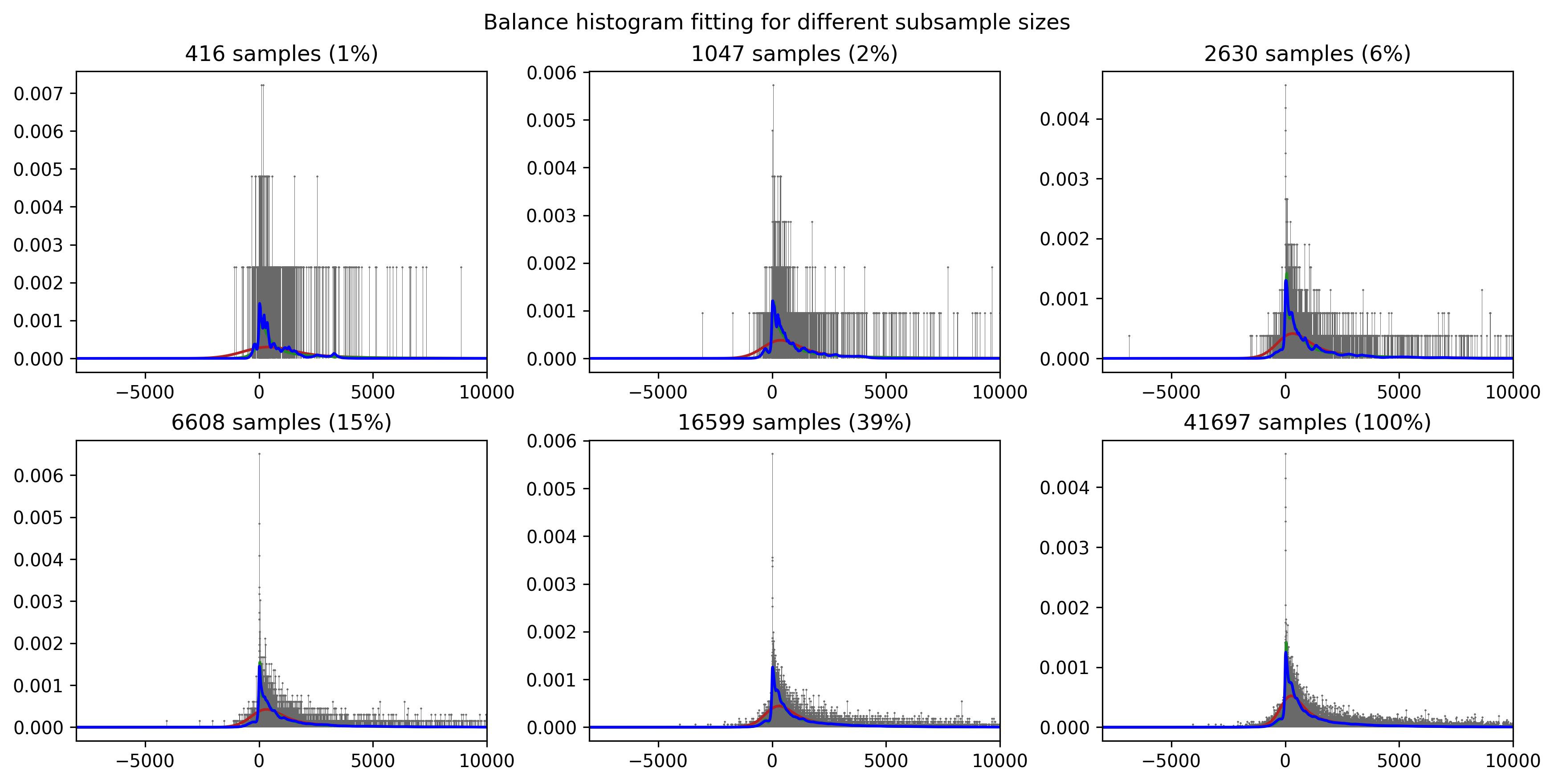}
    \caption{Empirical frequencies (gray) and PMF estimates for balance column of the bank marketing dataset. Gaussian kernel-density estimator in red, logpline in green, and our method in blue.}
    \label{fig:bank_balance}
\end{figure}

For the bank marketing data-set, the balance is already an integer, but can be negative. To apply our method, we first shifted the non-zero observations by the minimum balance, but plot the obtained PMF in the original coordinate system to make all methods comparable. In \figref{fig:bank_balance} we can see our method, logspline, and kernel-density estimation given various sample-sizes obtained by subsampling the balance column uniformly at random. Again, the kernel-density estimator struggles, whereas our method is competitive with, and almost identical to logspline.

\section{Summary, discussion, and future work}
The method we proposed in this paper is simple, fast, and reliable, with many design decisions made to make sure it is that way. Its main competitor, logspline, makes a special effort to insert spline knots (break-points of a piece-wise polynomial function) at the right place to model the changing curvature of the distribution, whereas our method is adaptive to the data by design through the construction of the appropriate eigenproblem. 

It is our hope that the idea of using eigen-problems to obtain a data-adaptive bases, by interpreting them as the standard \emph{fitting term} + \emph{regularizer} minimization problem, becomes useful to our readers in other domains. In fact, this was one the reasons for the way we chose to present the content of this paper.

The approach that is described here is not the only way we could devise a data-dependent basis, and reconstruct a PMF as a vector in the span of that basis. Below we describe several potential alternatives. A potential future work could be a more rigorous treatment of one (or more) of these and other alternatives. For each alternative, we also say why we have chosen \emph{not} chosen it.

\subsection{Alternative basis construction}
We interpreted the eigenvalue problem as a sequence of orthogonal vectors optimizing the balance between smoothness cost and alignment reward. The alignment reward can come in different forms, such as $\langle \vx^2, \vp^\alpha\rangle$ for some power $\alpha$. The choice of $\alpha=2$ may even appear natural, and also reduces to computing the eigenvectors of the tridiagonal matrix $\mL - \diag(\vp^\alpha)$, but we found that $\alpha=2$ performs badly, and believe that adding the complexity of tuning $\alpha$ not to be worthwhile. Alternatively, one can think of the reward $\langle \vp, \vx \rangle^2$, which reduces to computing the eigenvectors of $\mL - \vp \vp^T$. This time, it's not a tridiagonal matrix at all, and we do not have widespread reliable solvers for large-scale problems of this sort. It still has a special structure - the sum of a tridiagonal matrix and a matrix of rank 1, but developing dedicated eigensolvers is beyond the scope of this paper, even though such a formulation may turn useful.

The path laplacian matrix $\mL$ is also not the only way to express smoothness. Indeed, in \citet{semiclassical} the authors use a different, spectral differentiation matrix to express the smoothness penalty. However, such matrices are typically dense, and solving an eigenproblem of a large dense matrix is computationally expensive for large values of $N$.

Finally, constructing a data-dependent basis via a sequence of problems optimizing the balance between regularity and fitting can be generalized to other forms. For example, we can use the total-variation of $\vx$ as the regularizer, or the negative KL-divergence between $\vx$ and $\vp$ as the fitting reward. In addition, vector orthogonality is not the only way to express the notion of encoding information not present the solutions of the previous problems in the sequence. However, these are no longer simple eigenvalue problems, and obtaining such bases can be the subject of a paper in its own right.

\subsection{Alternatives to projection}
The work of \citet{semiclassical} on semi-classical signal analysis proposes a different way to reconstruct a smooth signal from noisy observations $\vp$. Instead of projection, the authors propose using the eigenvectors of $ \alpha^2 \mL - \diag(\vp)$ corresponding only to the negative eigenvalues $\lambda_1, \dots, \lambda_k$, where $\lambda_k$ is the largest negative eigenvalue, and estimate as
\[
\hat{p} = 4 \alpha \sum_{i=1}^k \sqrt{-\lambda_i} \vv_i^{2},
\]
where the square is taken component-wise. The tuned parameter is now $\alpha$, instead of the number of eigenvectors $k$.  Note, that $\vv_1^2, \dots, \vv_k^2$ are \emph{not} orthogonal, and the coefficients $\sqrt{-\lambda_i}$ are not projections. We have found this method to be less effective for noisy histograms, also in terms of reconstruction quality, but primarily since we have not found an effective way to tune $\alpha$ that is easy, fast, and reliable enough to be deployed at scale.

Moreover, instead of looking for vectors in the subspace spanned by the eigenspace, we could look for vectors in the intersection of this subspace in the unit simplex. Also, rather than the Euclidean distance, we could use any divergence measure $d$. The estimated PMF becomes $\vp^* = \mV \vz^*$, where $\vz^*$ is a solution of
\[
\min_{\vz} \quad d(\mV \vz, \vp) \quad \text{s.t.} \quad \mV \vz \in \Delta_N.
\]
When $d(\vx, \vy) = \| \vx - \vy\|_2$ we obtain the Euclidean projection onto the intersection of the subspace with the simplex\footnote{In the case of the eigenspace of $\mL - \diag(\vp)$, it can be shown that this intersection is always nonempty.}. But we can also have the KL-divergence $d(\vx, \vy) = \sum_{i} \evy_i \ln(\evy_i / \evx_i)$ as our divergence of choice, or any other divergence between two distributions. In many such cases, we obtain a simple convex optimization problem that is solvable by tools such as CVXPY \citep{cvxpy1,cvxpy2}.  While these alternatives are viable, and may yield better PMF estimates, it is not our goal in this work. Our goal is devising the simplest method we can that works reasonably well, and it is extremely hard to compete with the striking simplicity of just computing $\mV (\mV^T \vp)$ and re-normalizing. 

\bibliography{tmlr}
\bibliographystyle{tmlr}

\FloatBarrier
\appendix

\section{Estimator code snippet}\label{apx:estimator}
\begin{pycode}
def hist_estimator(samples):
    # compute empirical frequencies (assume support is [0 ... MAX])
    n = len(samples)
    p = np.bincount(samples, minlength=np.max(samples)).astype(float) / n

    # estimate upper bound on the number of eigenvectors - based on the logspline heuristic
    N = np.count_nonzero(p)
    K = int(math.ceil(min(4 * n ** (1/5), n / 4, N, 30)))

    # compute the first K eigenvectors
    diag = np.r_[1, np.full(p.size - 2, 2), 1] - p
    off_diag = np.full(p.size - 1, -1)
    _, V = scipy.linalg.eigh_tridiagonal(diag, off_diag, select="i", select_range=(0, k - 1))

    # estimate L2 error
    coefs = V.T @ p
    raw_2nd_moment = np.square(V).T @ p
    true_coef_sq_est = (
        np.maximum(0, n / (n - 1) * np.square(coefs) - 1 / (n - 1) * raw_2nd_moment) if n > 1
        else np.zeros_like(coefs)
    )
    err_estimate = (
        np.cumsum(raw_2nd_moment - true_coef_sq_est) / n +
        (np.cumsum(true_coef_sq_est[::-1])[::-1] - true_coef_sq_est)
    )

    # use k_best eigenvectors, and normalize
    k_best = np.argmin(risk_estimate) + 1
    u = V[:, :k_best] @ coefs[:k_best]
    return np.maximum(0, u) / np.sum(np.maximum(0,u))
\end{pycode}

\end{document}

%% file: math_commands.tex
\usepackage{amsmath,amsfonts,bm}

\def\figref#1{figure~\ref{#1}}
\def\Figref#1{Figure~\ref{#1}}

\def\Secref#1{Section~\ref{#1}}

\def\eqref#1{equation~\ref{#1}}

\def\1{\bm{1}}

\def\vc{{\bm{c}}}

\def\ve{{\bm{e}}}

\def\vp{{\bm{p}}}
\def\vq{{\bm{q}}}

\def\vs{{\bm{s}}}

\def\vu{{\bm{u}}}
\def\vv{{\bm{v}}}

\def\vx{{\bm{x}}}
\def\vy{{\bm{y}}}
\def\vz{{\bm{z}}}

\def\evp{{p}}
\def\evq{{q}}

\def\evx{{x}}
\def\evy{{y}}

\def\mA{{\bm{A}}}

\def\mH{{\bm{H}}}

\def\mL{{\bm{L}}}

\def\mV{{\bm{V}}}

\DeclareMathAlphabet{\mathsfit}{\encodingdefault}{\sfdefault}{m}{sl}
\SetMathAlphabet{\mathsfit}{bold}{\encodingdefault}{\sfdefault}{bx}{n}

\def\sN{{\mathbb{N}}}

\DeclareMathOperator*{\argmin}{arg\,min}